\documentclass[11pt,a4paper]{article}
\usepackage[hyperref]{acl2019}
\usepackage{times}
\usepackage{latexsym}

\usepackage{url}

\usepackage{enumitem}
\usepackage{graphicx}
\usepackage[labelformat=simple]{subcaption}

\usepackage{amsmath}
\usepackage{bm}
\usepackage[tone]{tipa}
\usepackage{cleveref}
\crefname{section}{§}{§§}
\Crefname{section}{Section}{}
\Crefname{figure}{Figure}{}
\Crefname{algorithm}{Algorithm}{}
\Crefname{equation}{Equation}{}
\usepackage[ruled]{algorithm2e}
\usepackage{xcolor}
\definecolor{cayenne}{RGB}{148,17,0}
\definecolor{moss}{RGB}{0,148,81}
\usepackage{booktabs}
\usepackage{multirow}
\newcommand{\specialcell}[2][c]{%
  \begin{tabular}[#1]{@{}c@{}}#2\end{tabular}}
\usepackage{array}
\newcolumntype{L}[1]{>{\raggedright\let\newline\\\arraybackslash\hspace{0pt}}m{#1}}
\newcolumntype{C}[1]{>{\centering\let\newline\\\arraybackslash\hspace{0pt}}m{#1}}
\newcolumntype{R}[1]{>{\raggedleft\let\newline\\\arraybackslash\hspace{0pt}}m{#1}}
\DeclareMathOperator*{\argmax}{arg\,max}

\aclfinalcopy 
\def\aclpaperid{1502} 

\title{AMR Parsing as Sequence-to-Graph Transduction}

\author{Sheng Zhang\qquad Xutai Ma\qquad Kevin Duh\qquad Benjamin Van Durme\\
  Johns Hopkins University\\
  \texttt{\{zsheng2, xutai\_ma\}@jhu.edu}\\
  \texttt{\{kevinduh, vandurme\}@cs.jhu.edu}}

\date{}

\begin{document}
\maketitle
\begin{abstract}
  We propose an attention-based model that treats AMR parsing as sequence-to-graph transduction.
  Unlike most AMR parsers that rely on pre-trained aligners, external semantic resources,
  or data augmentation, our proposed parser
  is aligner-free, and it can be effectively trained with limited amounts of labeled AMR data.
  Our experimental results outperform all previously reported {\sc Smatch} scores, on both AMR 2.0 (76.3\% F1 on LDC2017T10)
  and AMR 1.0 (70.2\% F1 on LDC2014T12).
\end{abstract}

\section{Introduction}

Abstract Meaning Representation (AMR, \citealp{AMR}) parsing is the task of transducing natural language text into AMR, a graph-based formalism used for capturing sentence-level semantics.
%
%
%
Challenges in AMR parsing include: (1) its property of reentrancy -- the same concept can participate in multiple relations --
which leads to graphs in contrast to trees~\citep{camr}; 
(2) the lack of gold alignments between nodes (concepts) in the graph and words in the text 
which limits attempts to rely on explicit alignments to generate training 
data~\citep{jamr,camr,E17-1051,P17-1043,E17-1035,P18-1170,D18-1198}; and
(3) relatively limited amounts of labeled data~\citep{P17-1014}.  

\begin{figure}[!t]
\centering
\includegraphics[width=0.47\textwidth]{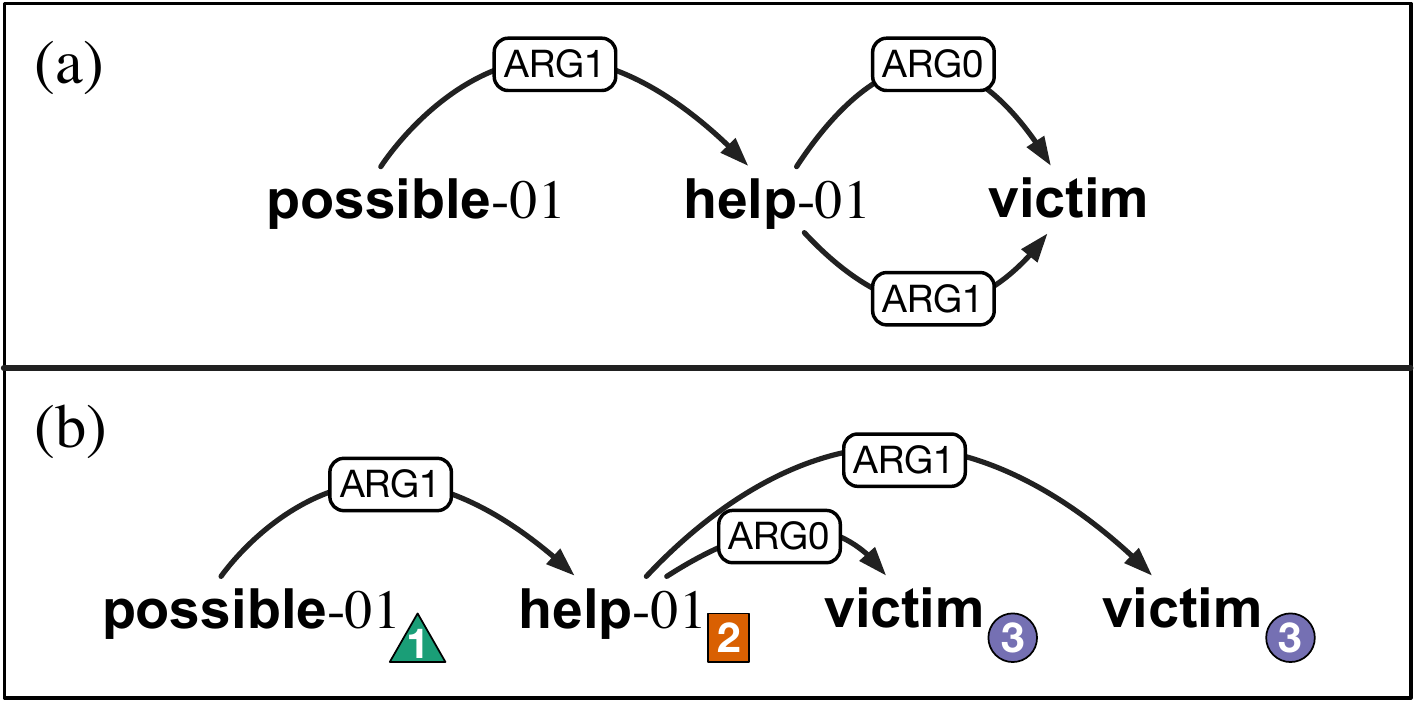}
\caption{Two views of reentrancy in AMR for an example sentence ``\emph{The victim could help himself.}''
(a) A standard AMR graph. (b) An AMR tree with node indices as an extra layer of annotation, 
where the corresponding graph can be recovered by merging nodes of the same index and unioning their incoming edges.
\label{fig:amr-ex}}
\end{figure}

Recent attempts to overcome these challenges include: modeling alignments as latent variables~\citep{P18-1037};
leveraging external semantic resources~\citep{D15-1198,S16-1182}; data augmentation~\citep{P17-1014,amr-seq2seq}; and employing attention-based sequence-to-sequence models~\cite{S16-1176,P17-1014,amr-seq2seq}.

In this paper, we introduce a different way to handle reentrancy, 
and propose an attention-based model that treats AMR parsing as sequence-to-graph transduction.
The proposed model, supported by an extended pointer-generator network, 
is aligner-free and can be effectively trained with limited amount of labeled AMR data.
Experiments on two publicly available AMR benchmarks demonstrate that 
our parser clearly outperforms the previous best parsers on both benchmarks.
It achieves the best reported \textsc{Smatch} scores: 76.3\% F1 on LDC2017T10 and 70.2\% F1 on LDC2014T12.
We also provide extensive ablative and qualitative studies, quantifying the contributions
from each component.
Our model implementation is available at \url{https://github.com/sheng-z/stog}.

\section{Another View of Reentrancy}
\label{sec:reentrancy}
AMR is a rooted, directed, and usually acyclic graph where nodes represent concepts, 
and labeled directed edges represent the relationships between them (see \Cref{fig:amr-ex} for an AMR example).
The reason for AMR being a graph instead of a tree is that it allows reentrant semantic relations.
For instance, in \Cref{fig:amr-ex}(a) ``\textbf{victim}'' is both \texttt{ARG0} and \texttt{ARG1} of ``\textbf{help}-01''. 
While efforts have gone into developing graph-based algorithms for AMR parsing~\citep{P13-1091,jamr},
it is more challenging to parse a sentence
into an AMR graph rather than a tree as there are efficient off-the-shelf tree-based algorithms, e.g., \citet{chu-liu-1965,edmonds1968optimum}.
To leverage these tree-based algorithms
as well as other structured prediction paradigms~\cite{P05-1012},
we introduce another view of reentrancy.

AMR reentrancy is employed when a node participates in multiple semantic relations.
We convert an AMR graph into a tree by duplicating nodes that have reentrant relations; that is,
whenever a node has a reentrant relation, we make a copy of that node 
and use the copy to participate in the relation, thereby resulting in a tree.
Next, in order to preserve the reentrancy information, we add an extra layer of annotation by
assigning an index to each node. 
Duplicated nodes are assigned the same index as the original node. 
\Cref{fig:amr-ex}(b) shows a resultant AMR tree:
subscripts of nodes are indices;
two ``\textbf{victim}'' nodes have the same index as they refer to the same concept.
The original AMR graph can be recovered by merging identically indexed nodes and unioning edges from/to these nodes.
Similar ideas were used by \citet{D15-1198} who introduced Skolem IDs to represent anaphoric references in the transformation from CCG to AMR,
and \citet{van-noord-bos-2017-dealing} who kept co-indexed AMR variables, and converted them to numbers.

\section{Task Formalization}
\label{sec:tech-overview}
\begin{figure}[!ht]
\centering
\includegraphics[width=0.36\textwidth]{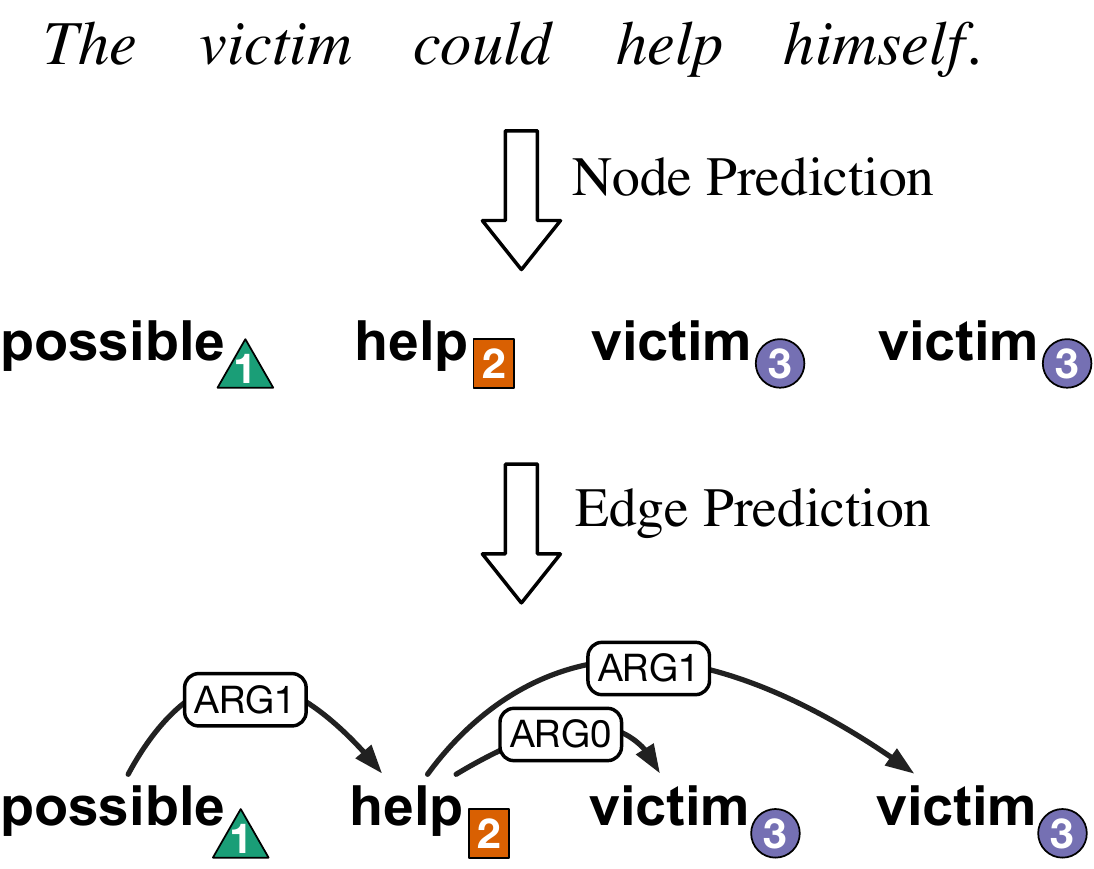}
\caption{A two-stage process of AMR parsing. 
We remove senses (i.e., -01, -02, etc.) as they will be assigned in the post-processing step.\label{fig:amr-task}} 
\end{figure}

If we consider the AMR tree with indexed nodes as the prediction target, then our approach to parsing is  formalized as a two-stage process: 
\textbf{node prediction} and \textbf{edge prediction}.\footnote{
The two-stage process is similar to ``\emph{concept identification}'' and ``\emph{relation identification}'' 
in \citet{jamr,D16-1065,P18-1037}; inter alia.}
An example of the parsing process is shown in \Cref{fig:amr-task}.

\noindent\textbf{Node Prediction} 
Given a input sentence $\bm{w}=\langle w_1, ..., w_n\rangle$, 
each $w_i$ a word in the sentence,
our approach \emph{sequentially} decodes a list of nodes $\bm{u}=\langle u_1, ..., u_m \rangle$ 
and \emph{deterministically} assigns their indices $\bm{d}=\langle d_1, ..., d_m\rangle$.
\begin{equation*}
    P(\bm{u})=\prod_{i=1}^mP(u_i\mid u_{<i}, d_{<i}, \bm{w})
\end{equation*}
Note that we allow the same node to occur multiple times in the list;
multiple occurrences of a node will be assigned the same index.
We choose to predict nodes sequentially rather than simultaneously,
because (1) we believe the current node generation is informative to   
the future node generation; 
(2) variants of efficient sequence-to-sequence models~\cite{bahdanau2014neural,vinyals2015grammar}
can be employed to model this process.
At the training time, we obtain the reference list of nodes and their indices 
using a pre-order traversal over the reference AMR tree. 
We also evaluate other traversal strategies, and will discuss their difference in \Cref{sec:results}.

\begin{figure*}[!t]
\centering
\includegraphics[width=0.95\textwidth]{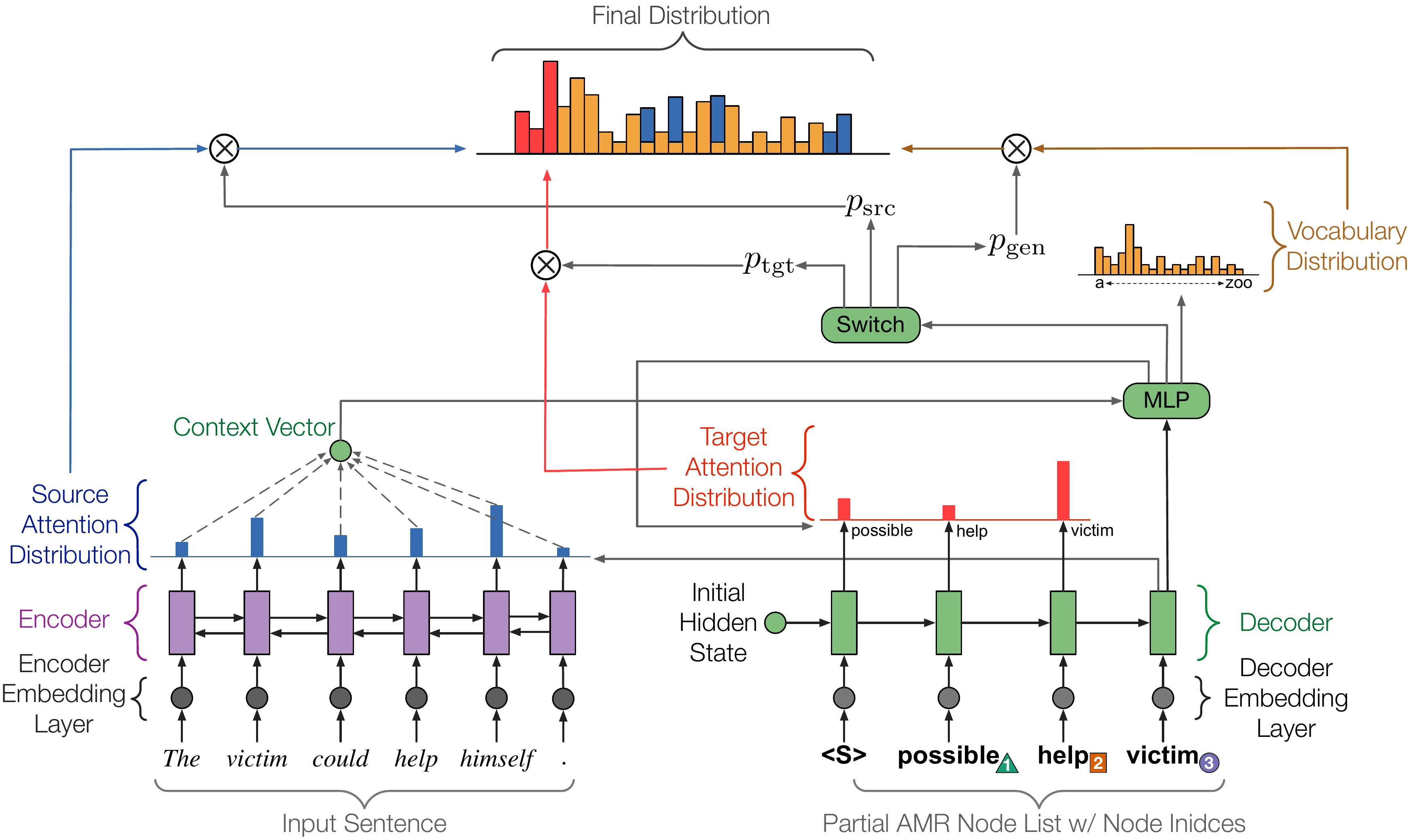}
\caption{Extended pointer-generator network for node prediction.
For each decoding time step, three probabilities $p_\textrm{src}$, $p_\textrm{tgt}$
and $p_\textrm{gen}$ are calculated. The source and target attention distributions as well as
the vocabulary distribution are weighted by these probabilities respectively, and then summed to obtain
the final distribution, from which we make our prediction. Best viewed in color.
\label{fig:pointer-generator}}
\end{figure*}

\noindent\textbf{Edge Prediction}
Given a input sentence $\bm{w}$, a node list $\bm{u}$, and indices $\bm{d}$, 
we look for the highest scoring parse tree $y$ in the space $\mathcal{Y}(\bm{u})$ of valid trees 
over $\bm{u}$ with the constraint of $\bm{d}$.
A parse tree $y$ is a set of directed head-modifier edges $y=\{(u_i, u_j)\mid 1\leq i,j\leq m\}$.
In order to make the search tractable,
we follow the arc-factored graph-based approach~\cite{P05-1012,Q16-1023}, decomposing the score of a tree
to the sum of the score of its head-modifier edges:
\begin{equation*}
    \textrm{parse}(\bm{u}) = \argmax_{y\in\mathcal{Y}(\bm{u})}\sum_{(u_i,u_j)\in y}\textrm{score}(u_i, u_j) 
\end{equation*}

Based on the scores of the edges, the highest scoring parse tree 
(i.e., maximum spanning arborescence) can be efficiently found using
the Chu-Liu-Edmonnds algorithm.
We further incorporate indices as constraints in the algorithm, which is described in \Cref{sec:prediction}. 
After obtaining the parse tree, we merge identically indexed nodes to recover the standard AMR graph. 

\section{Model}
Our model has two main modules:
(1) an extended pointer-generator network for node prediction;
and (2) a deep biaffine classifier for edge prediction.
The two modules correspond to the two-stage process for AMR parsing,
and they are \emph{jointly} learned during training.

\subsection{Extended Pointer-Generator Network}
Inspired by the \emph{self-copy} mechanism in \citet{x-dsp},
we extend the pointer-generator network~\cite{pointer-generator} for node prediction.
The pointer-generator network was proposed for text summarization,
which can copy words from the source text via \emph{pointing}, while retaining the 
ability to produce novel words through the \emph{generator}.
The major difference of our extension is that it can copy nodes, 
not only from the source text,
but also from the previously generated nodes on the target side. 
This \emph{target-side pointing} is well-suited to our task as
nodes we will predict can be copies of other nodes.
While there are other pointer/copy networks~\cite{pointer1,merity2016pointer,pointer2,pointer3,pointer4},
we found the pointer-generator network very effective at reducing data sparsity in AMR parsing,
which will be shown in \Cref{sec:results}.

As depicted in \Cref{fig:pointer-generator},
the extended pointer-generator network consists of four major components:
an encoder embedding layer, an encoder, a decoder embedding layer, and a decoder.

\noindent\textbf{Encoder Embedding Layer}
This layer converts words in input sentences into vector representations. 
Each vector is the concatenation of embeddings of GloVe~\cite{glove}, 
BERT~\cite{devlin2018bert}, POS (part-of-speech) tags and anonymization indicators,
and features learned by a character-level convolutional neural network (CharCNN,~\citealp{charCNN}).

Anonymization indicators are binary indicators that tell the encoder whether 
the word is an anonymized word. In preprocessing,
text spans of named entities in input sentences will be replaced by anonymized tokens
(e.g. \texttt{person}, \texttt{country}) to reduce sparsity~(see the Appendix for details).

\begin{figure}[!ht]
\centering
\includegraphics[width=0.35\textwidth]{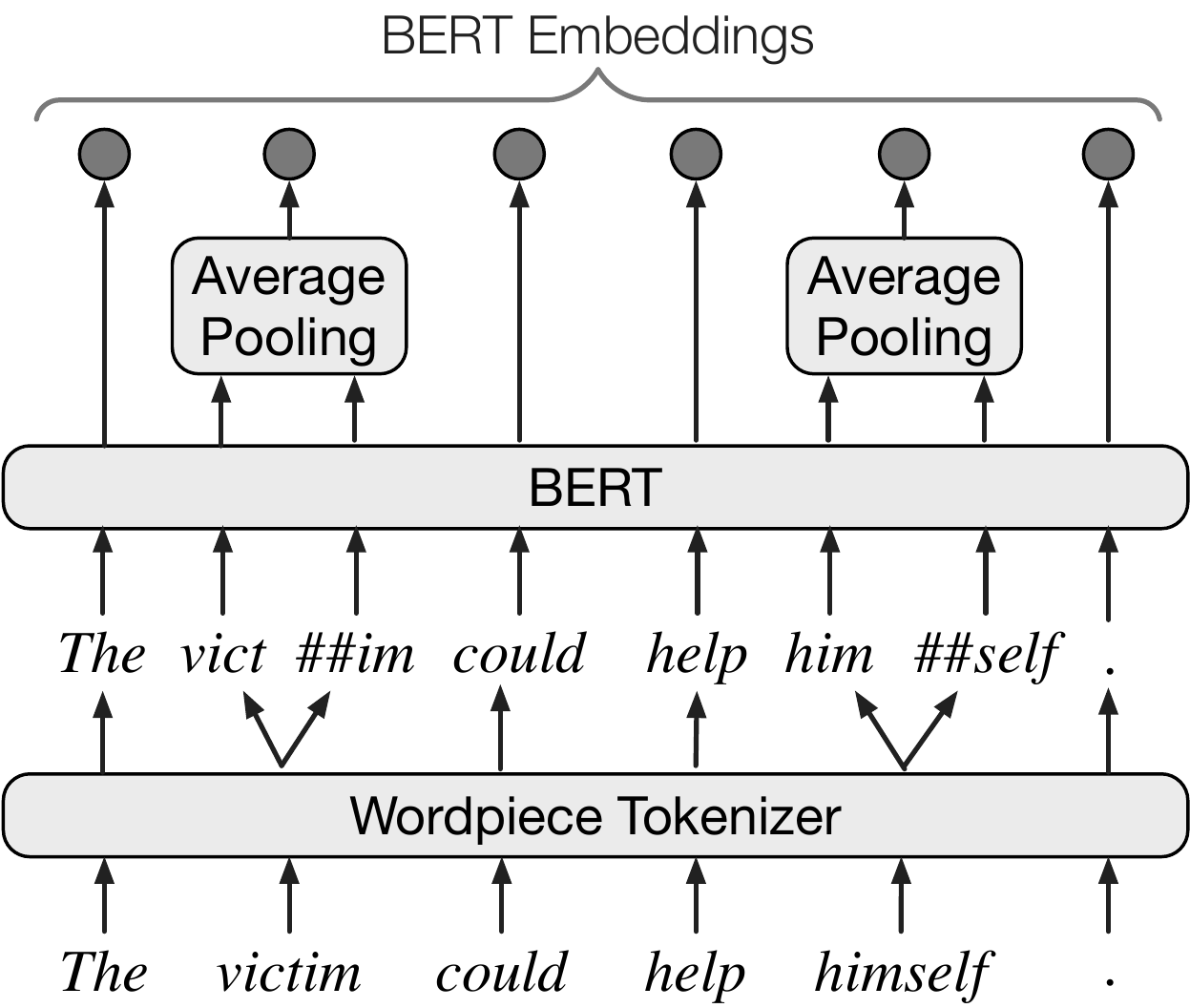}
\caption{Word-level embeddings from BERT.\label{fig:bert-encoder}}
\end{figure}

Except BERT, all other embeddings are fetched from their corresponding learned embedding look-up tables.
BERT takes subword units as input,
which means that one word may correspond to multiple hidden states of BERT.
In order to accurately use these hidden states to represent each word,
we apply an average pooling function to the outputs of BERT.
\Cref{fig:bert-encoder} illustrates the process of generating word-level embeddings from BERT.

\noindent\textbf{Encoder}
The encoder is a multi-layer bidirectional RNN~\cite{schuster1997bidirectional}:
\begin{equation*}
\bm{h}^l_i = [\overrightarrow{f}^l(\bm{h}^{l-1}_i, \bm{h}^l_{i-1});\overleftarrow{f}^l(\bm{h}^{l-1}_i, \bm{h}^l_{i+1})],
\end{equation*}
where $\overrightarrow{f}^l$ and $\overleftarrow{f}^l$ are two LSTM cells~\cite{hochreiter1997long}; 
$\bm{h}^l_i$ is the $l$-th layer encoder hidden state at the time step $i$;
$\bm{h}^0_i$ is the encoder embedding layer output for word $w_i$.

\noindent\textbf{Decoder Embedding Layer}
Similar to the encoder embedding layer, this layer outputs vector representations for 
AMR nodes. The difference is that each vector is the concatenation of embeddings of 
GloVe, POS tags and indices, and feature vectors from CharCNN.

POS tags of nodes are inferred at runtime: 
if a node is a copy from the input sentence, the POS tag of the corresponding word is used;
if a node is a copy from the preceding nodes, the POS tag of its antecedent is used;
if a node is a new node emitted from the vocabulary, an \texttt{UNK} tag is used.

We do not include BERT embeddings in this layer
because AMR nodes, especially their order, are 
significantly different from natural language text (on which BERT was pre-trained).
We tried to use ``fixed'' BERT in this layer, which did not lead to improvement.\footnote{
Limited by the GPU memory, we do not fine-tune BERT on this task and leave it for future work.}

\noindent\textbf{Decoder}
At each step $t$, the decoder (an $l$-layer unidirectional LSTM) receives
hidden state $\bm{s}^{l-1}_t$ from the last layer and 
hidden state $\bm{s}^l_{t-1}$ from the previous time step, and generates hidden state $\bm{s}^l_t$: 
\begin{equation*}
    \bm{s}^l_t = f^l(\bm{s}^{l-1}_t, \bm{s}^l_{t-1}),
\end{equation*}
where $\bm{s}^0_t$ is the concatenation (i.e., the \emph{input-feeding} approach, \citealp{D15-1166}) of two vectors:
the decoder embedding layer output for the previous node $u_{t-1}$
(while training, $u_{t-1}$ is the previous node of the reference node list;
at test time it is the previous node emitted by the decoder),
and the attentional vector $\widetilde{\bm{s}}_{t-1}$ from the previous step (explained later in this section).
$\bm{s}^l_0$ is the concatenation of last \emph{encoder hidden states} 
from $\overrightarrow{f}^l$ and $\overleftarrow{f}^l$ respectively.

\emph{Source attention distribution} $\bm{a}^t_\textrm{src}$ is calculated
by additive attention~\cite{bahdanau2014neural}:
\begin{align*}
    \bm{e}^t_\textrm{src} = &~\bm{v}^{\top}_\textrm{src}\textrm{tanh}(\bm{W}_\textrm{src}\bm{h}^l_{1:n} + \bm{U}_\textrm{src}\bm{s}^l_t + \bm{b}_\textrm{src}), \\
    \bm{a}^t_\textrm{src} = &~\textrm{softmax}(\bm{e}^t_\textrm{src}),
\end{align*}
and it is then used to produce a weighted sum of encoder hidden states, i.e., the context vector $\bm{c}_t$.

\emph{Attentional vector} $\widetilde{\bm{s}}_t$ combines both source and target side information, and it is calculated by 
an MLP (shown in \Cref{fig:pointer-generator}):
\begin{equation*}
    \widetilde{\bm{s}}_t = \textrm{tanh}(\bm{W}_c[\bm{c}_t;\bm{s}^l_t] + \bm{b}_c)
\end{equation*}
The attentional vector $\widetilde{\bm{s}}_t$ has 3 usages:

\noindent(1) it is fed through a linear layer and softmax to produce the vocabulary distribution:
\begin{equation*}
    P_\textrm{vocab} = \textrm{softmax}(\bm{W}_\textrm{vocab}\widetilde{\bm{s}}_t + \bm{b}_\textrm{vocab})
\end{equation*}

\noindent(2) it is used to calculate the \emph{target attention distribution} $\bm{a}^t_\textrm{tgt}$:
\begin{align*}
    \bm{e}^t_\textrm{tgt} = &~\bm{v}^{\top}_\textrm{tgt}\textrm{tanh}(\bm{W}_\textrm{tgt}\widetilde{\bm{s}}_{1:t-1} + \bm{U}_\textrm{tgt}\widetilde{\bm{s}}_t + \bm{b}_\textrm{tgt}), \\
    \bm{a}^t_\textrm{tgt} = &~\textrm{softmax}(\bm{e}^t_\textrm{tgt}),
\end{align*}

\noindent(3) it is used to calculate \emph{source-side copy} probability $p_\textrm{src}$,
\emph{target-side copy} probability $p_\textrm{tgt}$, and 
\emph{generation} probability $p_\textrm{gen}$ via a \emph{switch} layer:
\begin{equation*}
    [p_\textrm{src},p_\textrm{tgt},p_\textrm{gen}] = \textrm{softmax}(\bm{W}_\textrm{switch}\widetilde{\bm{s}}_t + \bm{b}_\textrm{switch})
\end{equation*}
Note that $p_\textrm{src}+p_\textrm{tgt}+p_\textrm{gen}=1$.
They act as a soft switch to choose between 
\emph{copying} an existing node from the preceding nodes by
sampling from the target attention distribution $\bm{a}^t_\textrm{tgt}$,
or \emph{emitting} a new node in two ways: 
(1) \emph{generating} a new node from the fixed vocabulary by sampling from $P_\textrm{vocab}$, or
(2) \emph{copying} a word (as a new node) from the input sentence by sampling 
from the source attention distribution $\bm{a}^t_\textrm{src}$.

The \emph{final probability distribution} $P^\textrm{(node)}(u_t)$ for node $u_t$ is defined as follows. If $u_t$ is a copy of existing nodes, then:
\begin{equation*}
    P^\textrm{(node)}(u_t) = p_\textrm{tgt}\sum_{i:u_i=u_t}^{t-1}\bm{a}^t_\textrm{tgt}[i],
\end{equation*}
otherwise:
\begin{equation*}
    P^\textrm{(node)}(u_t) =
    p_\textrm{gen}P_\textrm{vocab}(u_t) +
    p_\textrm{src}\sum_{i:w_i=u_t}^{n}\bm{a}^t_\textrm{src}[i],
\end{equation*}
where $\bm{a}^t[i]$ indexes the $i$-th element of $\bm{a}^t$.
Note that a new node may have the same surface form as the existing
node. We track their difference using indices.
The index $d_t$ for node $u_t$ is assigned \emph{deterministically} as below:
\begin{align*}
    d_t = \left \{
        \begin{aligned}
            t, &\ \textrm{if}\ u_t \textrm{ is a new node;}\\
            d_j, &\ \textrm{if}\ u_t \textrm{ is a copy of its antecedent } u_j\textrm{.}
        \end{aligned}
    \right.
\end{align*}

\subsection{Deep Biaffine Classifier}
For the second stage (i.e., edge prediction), we employ a deep biaffine classifier,
which was originally proposed for graph-based dependency parsing~\cite{dozat2016deep},
and recently has been applied to semantic parsing~\cite{P17-1186,P18-2077}.

\begin{figure}[!t]
\centering
\includegraphics[width=0.47\textwidth]{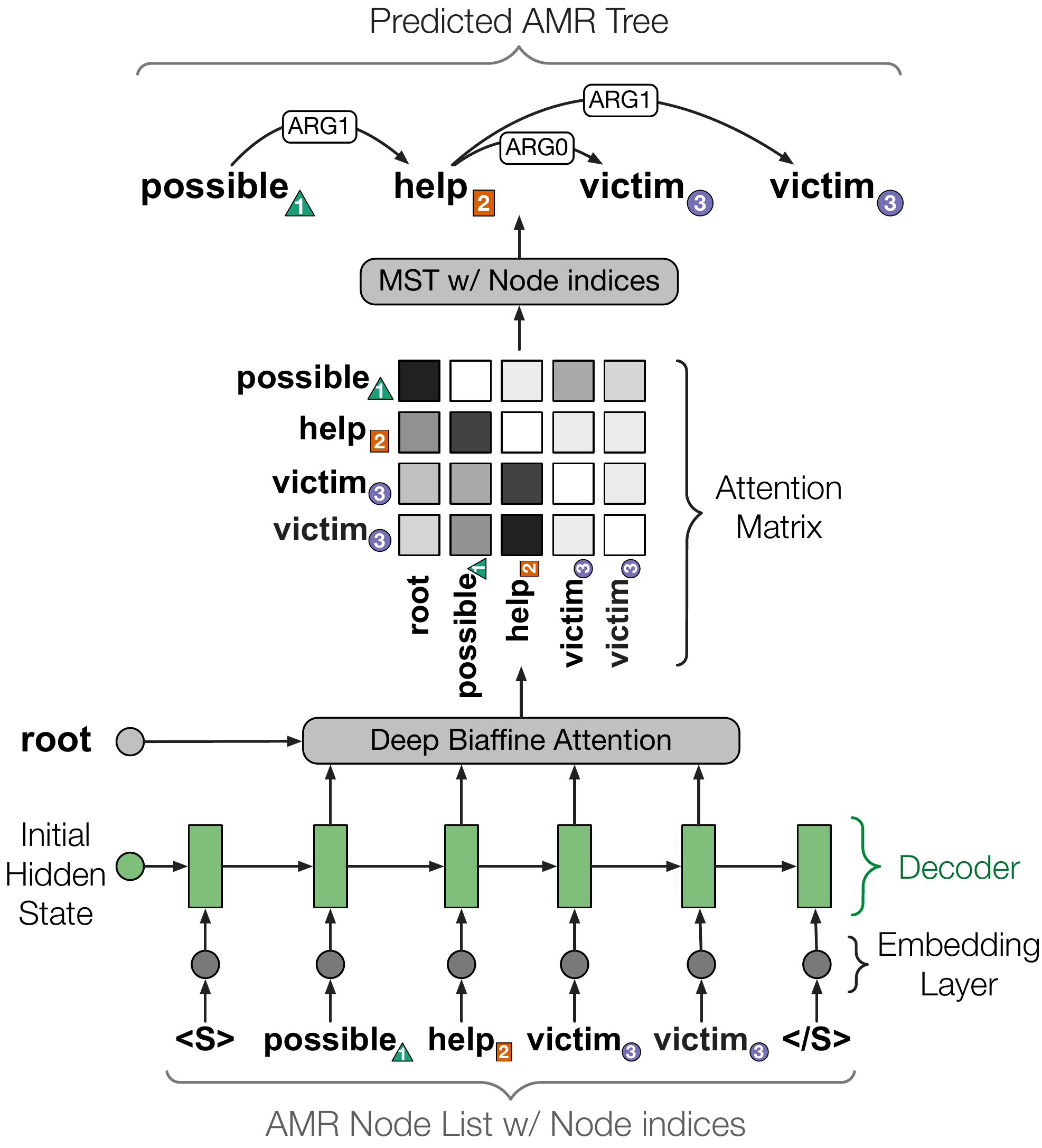}
\caption{Deep biaffine classifier for edge prediction. Edge label prediction is not depicted in the figure.\label{fig:biaffine-classifier}}
\end{figure}

As depicted in \Cref{fig:biaffine-classifier},
the major difference of our usage is that instead of re-encoding AMR nodes, we directly use
\emph{decoder hidden states} from the extended pointer-generator network as the input to deep biaffine classifier.
We find two advantages of using decoder hidden states as input: 
(1) through the \emph{input-feeding} approach, decoder hidden states contain
contextualized information from both the input sentence and the predicted nodes;
(2) because decoder hidden states are used for both node prediction and edge prediction,
we can jointly train the two modules in our model.

Given decoder hidden states $\langle\bm{s}_1,...,\bm{s}_m\rangle$ and 
a learnt vector representation $\bm{s}^\prime_0$ of a dummy root, 
we follow \citet{dozat2016deep}, factorizing edge prediction into two components:
one that predicts whether or not a directed edge $(u_k, u_t)$ exists between two nodes $u_k$ and $u_t$,
and another that predicts the best label for each potential edge.

Edge and label scores are calculated as below:
\begin{gather*}
    \bm{s}^\textrm{(edge-head)}_t = \textrm{MLP}^\textrm{(edge-head)}(\bm{s}_t)\\
    \bm{s}^\textrm{(edge-dep)}_t = \textrm{MLP}^\textrm{(edge-dep)}(\bm{s}_t) \\
    \bm{s}^\textrm{(label-head)}_t = \textrm{MLP}^\textrm{(label-head)}(\bm{s}_t)\\
    \bm{s}^\textrm{(label-dep)}_t = \textrm{MLP}^\textrm{(label-dep)}(\bm{s}_t) \\
    \textrm{score}^\textrm{(edge)}_{k,t} = \textrm{Biaffine}(\bm{s}^\textrm{(edge-head)}_k, \bm{s}^\textrm{(edge-dep)}_t)\\
    \textrm{score}^\textrm{(label)}_{k,t} = \textrm{Bilinear}(\bm{s}^\textrm{(label-head)}_k, \bm{s}^\textrm{(label-dep)}_t)
\end{gather*}
where MLP, Biaffine and Bilinear are defined as below:
\begin{gather*}
    \textrm{MLP}(\bm{x}) = \textrm{ELU}(\bm{W}\bm{x}+\bm{b})\\
    \textrm{Biaffine}(\bm{x}_1, \bm{x}_2) = \bm{x}^\top_1\bm{U}\bm{x}_2 + \bm{W}[\bm{x}_1;\bm{x}_2] + \bm{b} \\
    \textrm{Bilinear}(\bm{x}_1, \bm{x}_2) = \bm{x}^\top_1\bm{U}\bm{x}_2 + \bm{b} 
\end{gather*}

Given a node $u_t$, the probability of $u_k$ being the edge head of $u_t$ is defined as:
\begin{equation*}
    P^{\textrm{(head)}}_t(u_k)=\frac{\exp(\textrm{score}^\textrm{(edge)}_{k,t})}{\sum_{j=1}^m\exp(\textrm{score}^\textrm{(edge)}_{j,t})}
\end{equation*}

The edge label probability for edge $(u_k, u_t)$ is defined as:
\begin{equation*}
    P^\textrm{(label)}_{k,t}(l)=\frac{\exp(\textrm{score}^\textrm{(label)}_{k,t}[l])}{\sum_{l^\prime}\exp(\textrm{score}^\textrm{(label)}_{k,t}[l^\prime])}
\end{equation*}

\subsection{Training}
\label{sec:training}
The training objective is to jointly minimize the loss of reference nodes and edges,
which can be decomposed to the sum of the negative log likelihood at each time step $t$ for (1) the reference 
node $u_t$, (2) the reference edge head $u_k$ of node $u_t$, and (3) the reference edge label $l$ between $u_k$ and $u_t$:
\begin{align*}
    \textrm{minimize}-\sum_{t=1}^m & [\log{P^\textrm{(node)}(u_t)} + \log{P^{\textrm{(head)}}_t(u_k)} \\
    & + \log{P^\textrm{(label)}_{k,t}(l)} + \lambda\textrm{covloss}_t]
\end{align*}
$\textrm{covloss}_t$ is a \emph{coverage loss} to penalize repetitive nodes:
$\textrm{covloss}_t=\sum_i\textrm{min}(\bm{a}^t_\textrm{src}[i],\textbf{cov}^t[i])$, where
$\textbf{cov}^t$ is the sum of source attention
distributions over all previous decoding time steps:
$\textbf{cov}^t = \sum_{t^\prime=0}^{t-1}\bm{a}^{t^\prime}_\textrm{src}$.
See \citet{pointer-generator} for full details.

\subsection{Prediction}
\label{sec:prediction}
For node prediction, based on the final probability distribution $P^\textrm{(node)}(u_t)$ at each decoding time step, 
we implement both greedy search and beam search to sequentially decode a node list $\bm{u}$ and indices $\bm{d}$.

For edge prediction, given the predicted node list $\bm{u}$, 
their indices $\bm{d}$, and 
the edge scores $S=\{\textrm{score}^\textrm{(edge)}_{i,j} \mid 0\leq i,j\leq m\}$,
we apply the Chu-Liu-Edmonds algorithm with a simple adaption to 
find the maximum spanning tree (MST).
As described in Algorithm \ref{alg:mst},
before calling the Chu-Liu-Edmonds algorithm, 
we first include a dummy root $u_0$ to ensure every node have a head, 
and then exclude edges whose source and destination nodes have the same indices,
because these nodes will be merged into a single node to recover the standard AMR graph
where self-loops are invalid.

\begin{algorithm}[!ht]
\small
\SetAlCapNameFnt{\small}
\SetAlCapFnt{\small}
\SetKwFunction{MST}{MST}
\SetKwProg{Fn}{Function}{}{}
\SetKwInOut{Input}{Input}\SetKwInOut{Output}{Output}
\Input{Nodes $\bm{u}=\langle u_1, ...,u_m\rangle$,\\ 
Indices $\bm{d}=\langle d_1,...d_m\rangle$,\\
Edge scores $S=\{\textrm{score}^\textrm{(edge)}_{i,j}\mid 0\leq i,j\leq m\}$}
\Output{A maximum spanning tree.}
\tcp{Include the dummy root $u_0$.}
$V\leftarrow\{u_0\}\cup\bm{u}$\;
$d_0\leftarrow 0$\;
\BlankLine
\tcp{Exclude invalid edges.}
\tcp{$d_i$ is the node index for node $u_i$.}
$E\leftarrow\{(u_i, u_j)\mid 0\leq i,j\leq m; d_i\neq d_j\}$\;
\BlankLine
\tcp{Chu-Liu-Edmonds algorithm}
\Return \MST{$V, E, S, u_0$}\;
\caption{Chu-Liu-Edmonds algo. w/ Adaption}\label{alg:mst}
\end{algorithm}

\section{Related Work}
AMR parsing approaches can be categorized into \emph{alignment}-based,
\emph{transition}-based, \emph{grammar}-based, and \emph{attention}-based approaches.

Alignment-based approaches were first explored by JAMR~\cite{jamr}, a pipeline of 
concept and relation identification with a graph-based algorithm.
\citet{D16-1065} improved this by jointly learning concept and relation identification
with an incremental model. Both approaches rely on features based on alignments.
\citet{P18-1037} treated alignments as latent variables in a joint probabilistic model, leading to a substantial reported improvement. 
Our approach requires no explicit alignments, but implicitly learns a source-side copy
mechanism using attention.

Transition-based approaches began with \citet{camr,S16-1181}, who incrementally transform dependency parses into AMRs using transiton-based models, which was
followed by a line of research, such as \citet{S16-1178,S16-1179,S16-1180,E17-1051,ballesteros-al-onaizan-2017-amr,P18-1170}.
A pre-trained aligner, e.g. \citet{D14-1048,liu-etal-2018-amr}, is needed for most parsers to generate training 
data (e.g., oracles for a transition-based parser).
Our approach makes no significant use of external semantic resources,\footnote{
We only use POS tags in the core parsing task. 
In post-processing, we use an entity linker as a common move for wikification like \citet{amr-seq2seq}.
}
and is aligner-free.

Grammar-based approaches are represented by
\citet{D15-1198,peng-etal-2015-synchronous} who leveraged external semantic resources, and employed CCG-based or SHRG-based grammar induction approaches
converting logical forms into AMRs.
\citet{D15-1136} recast AMR parsing as a machine translation problem, while also drawing features from 
external semantic resources.

Attention-based parsing with Seq2Seq-style models have been considered \cite{S16-1176,E17-1035},
but are limited by the relatively small amount of labeled AMR data.
\citet{P17-1014} overcame this by making use of millions of unlabeled data through self-training, while
\citet{amr-seq2seq} showed significant gains via 
a character-level Seq2Seq model and a large amount of silver-standard AMR training data.
In contrast, our approach supported by extended pointer generator can 
be effectively trained on the limited amount of labeled AMR data, with no data augmentation.

\section{AMR Pre- and Post-processing}
\label{sec:preproc}
Anonymization is often used in AMR preprocessing 
to reduce sparsity~(\citealp{P15-1095,E17-1035,D18-1198}, inter alia).
Similar to~\citet{P17-1014},  we anonymize sub-graphs of named entities and other entities.
Like~\citet{P18-1037}, we remove senses, and use Stanford CoreNLP~\cite{manning-EtAl:2014:P14-5} 
to lemmatize input sentences and add POS tags.

In post-processing, we assign the most frequent sense for nodes (-01, if unseen) like~\citet{P18-1037},
and restore wiki links using the DBpedia Spotlight API~\cite{isem2013daiber} following \citet{S16-1182,amr-seq2seq}.
We add polarity attributes based on the rules observed from the training data.
More details of pre- and post-processing are provided in the Appendix.

\section{Experiments}

\subsection{Setup}
\begin{table}[!ht]
\small
\centering
\begin{tabular}{@{}ll@{}}
\toprule
\multicolumn{2}{l}{\textbf{GloVe.840B.300d embeddings}}             \\ 
~~~dim                           & 300                       \\\midrule
\multicolumn{2}{l}{\textbf{BERT embeddings}}              \\
~~~source                        & BERT-Large-cased          \\
~~~dim                           & 1024                      \\\midrule
\multicolumn{2}{l}{\textbf{POS tag embeddings}}           \\
~~~dim                           & 100                       \\\midrule
\multicolumn{2}{l}{\textbf{Anonymization indicator embeddings}} \\
~~~dim                           & 50                        \\\midrule
\multicolumn{2}{l}{\textbf{Index embeddings}}             \\
~~~dim                           & 50                        \\\midrule
\multicolumn{2}{l}{\textbf{CharCNN}}                      \\
~~~num\_filters                  & 100                       \\
~~~ngram\_filter\_sizes          & {[}3{]}                   \\\midrule
\multicolumn{2}{l}{\textbf{Encoder}}                      \\
~~~hidden\_size                  & 512                       \\
~~~num\_layers                   & 2                         \\\midrule
\multicolumn{2}{l}{\textbf{Decoder}}                      \\
~~~hidden\_size                  & 1024                      \\
~~~num\_layers                   & 2                         \\\midrule
\multicolumn{2}{l}{\textbf{Deep biaffine classifier}}     \\
~~~edge\_hidden\_size            & 256                       \\
~~~label\_hidden\_size           & 128                       \\\midrule
\multicolumn{2}{l}{\textbf{Optimizer}}                    \\
~~~type                          & ADAM                      \\
~~~learning\_rate                & 0.001                     \\
~~~max\_grad\_norm               & 5.0                       \\ \midrule
~~~\textbf{Coverage loss weight} $\lambda$                             & 1.0                       \\ \midrule
~~~\textbf{Beam size}                             & 5                       \\ \midrule
\multicolumn{2}{l}{\textbf{Vocabulary}}                   \\
~~~encoder\_vocab\_size (AMR 2.0)          & 18000                     \\
~~~decoder\_vocab\_size (AMR 2.0)         & 12200                     \\ 
~~~encoder\_vocab\_size (AMR 1.0)         & 9200                     \\
~~~decoder\_vocab\_size (AMR 1.0)         & 7300                     \\ \midrule
~~~\textbf{Batch size}           & 64                        \\ \bottomrule
\end{tabular}
\caption{Hyper-parameter settings}
\label{tab:hyper}
\end{table}

We conduct experiments on two AMR general releases (available to all LDC subscribers): 
AMR 2.0 (LDC2017T10) and AMR 1.0 (LDC2014T12).
Our model is trained using ADAM~\cite{kingma2014adam} for up to 120 epochs, with early stopping based on the development set.
Full model training takes about 19 hours on AMR 2.0 and 7 hours on AMR 1.0, using two GeForce GTX TITAN X GPUs.
At training, we have to fix BERT parameters due to the limited GPU memory.
We leave fine-tuning BERT for future work.

\Cref{tab:hyper} lists the hyper-parameters used in our full model.
Both encoder and decoder embedding layers have GloVe and POS tag embeddings as well as CharCNN,
but their parameters are not tied.
We apply dropout (dropout\_rate = 0.33) to the outputs of each module.

\subsection{Results}
\label{sec:results}

\begin{table}[!ht]
\centering
\begin{tabular}{@{}cll@{}}
\toprule
Corpus                     & Parser & \multicolumn{1}{c}{F1(\%)} \\ \midrule
\multirow{5}{*}{\specialcell{AMR\\ 2.0}} & \citet{S17-2157}                           & 61.9                  \\
                         & \citet{amr-seq2seq}                           & 71.0$^*$                  \\
                         & \citet{P18-1170}                           & 71.0$\pm$0.5                  \\
                         & \citet{P18-1037}                           & 74.4$\pm$0.2                  \\ 
                         & \citet{naseem-etal-2019-amr}                           & 75.5                  \\ \cmidrule(l){2-3}
                         & Ours                       & \textbf{76.3}$\pm$0.1                  \\ \midrule
\multirow{5}{*}{\specialcell{AMR\\ 1.0}} & \citet{S16-1186}                            & 66.0                  \\
                         &   \citet{D15-1136}                        & 67.1                  \\
                         &   \citet{D17-1129}                         & 68.1                  \\
                         &   \citet{D18-1198}                         & 68.3$\pm$0.4                  \\ \cmidrule(l){2-3}
                         & Ours                       & \textbf{70.2}$\pm$0.1                 \\ \bottomrule
\end{tabular}
\caption{\textsc{Smatch} scores on the test sets of AMR 2.0 and 1.0. Standard deviation is computed over 3 runs with different random seeds.
$^*$ indicates the previous best score from attention-based models.}
\label{tab:main-results}
\end{table}

\noindent\textbf{Main Results}
We compare our approach against the previous best approaches and 
several recent competitors.
\Cref{tab:main-results} summarizes their \textsc{Smatch} scores~\cite{P13-2131} on the test sets of two AMR general releases.
On AMR 2.0, we outperform the latest push from \citet{naseem-etal-2019-amr} by 0.8\% F1, and significantly improves \citet{P18-1037}'s results by 1.9\% F1.
Compared to the previous best attention-based approach~\cite{amr-seq2seq}, our approach shows a substantial gain of 5.3\% F1, with no usage of any silver-standard training data.
On AMR 1.0 where the traininng instances are only around 10k, 
we improve the best reported results by 1.9\% F1.

\noindent\textbf{Fine-grained Results}
In \Cref{tab:individual-phenom}, we assess the quality of each subtask
using the AMR-evaluation tools~\cite{E17-1051}. 
We see a notable increase on reentrancies, which we attribute to target-side copy 
(based on our ablation studies in the next section).
Significant increases are also shown on wikification and negation, indicating 
the benefits of using DBpedia Spotlight API and negation detection rules in post-processing.
On all other subtasks except named entities, our approach achieves competitive results to the previous
best approaches~\cite{P18-1037,naseem-etal-2019-amr}, and outperforms the previous best attention-based approach~\cite{amr-seq2seq}.
The difference of scores on named entities is mainly caused by  
anonymization methods used in preprocessing,
which suggests a potential improvement by adapting the anonymization method presented in \citet{P18-1037}
to our approach.

\begin{table}[!t]
\centering
\begin{tabular}{@{}lcccl@{}}
\toprule
Metric       & vN'18 & L'18 & N'19 & \multicolumn{1}{c}{Ours} \\ \midrule
\textsc{Smatch}  & 71.0   & 74.4 & 75.5   & \textbf{76.3}$\pm$0.1 \\ \midrule
Unlabeled    & 74   & 77  & \textbf{80}  & 79.0$\pm$0.1 \\
No WSD       & 72   & 76  & 76  & \textbf{76.8}$\pm$0.1 \\
Reentrancies & 52   & 52  & 56  & \textbf{60.0}$\pm$0.1 \\
Concepts     & 82   & \textbf{86} & \textbf{86}   & 84.8$\pm$0.1 \\
Named Ent.   & 79   & \textbf{86}   & 83 &  77.9$\pm$0.2 \\
Wikification & 65   & 76  & 80  & \textbf{85.8}$\pm$0.3 \\
Negation     & 62   & 58  & 67  & \textbf{75.2}$\pm$0.2 \\
SRL          & 66   & 70  & \textbf{72}  & 69.7$\pm$0.2 \\ \bottomrule
\end{tabular}
\caption{Fine-grained F1 scores on the AMR 2.0 test set. vN'17 is \citet{amr-seq2seq};
L'18 is \citet{P18-1037}; N'19 is \citet{naseem-etal-2019-amr}.}
\label{tab:individual-phenom}
\end{table}

\begin{table}[!ht]
\centering
\begin{tabular}{@{}lcc@{}}
\toprule
Ablation & \specialcell{AMR\\ 1.0} & \specialcell{AMR\\ 2.0} \\ \midrule
Full model                   & 70.2                        & 76.3                        \\ \midrule
no source-side copy          & 62.7                        &   70.9                     \\
no target-side copy          & 66.2                       &     71.6                   \\
no coverage loss             & 68.5                        & 74.5                        \\
no BERT embeddings           & 68.8                        & 74.6                        \\
no index embeddings     & 68.5                        & 75.5                        \\ 
no anonym. indicator embed. & 68.9                        & 75.6                        \\
no beam search               & 69.2                        & 75.3                        \\
no POS tag embeddings        & 69.2                        & 75.7                        \\
no CharCNN features                  & 70.0                        & 75.8                        \\ \midrule
only edge prediction & 88.4                        & 90.9                        \\ \bottomrule
\end{tabular}
\caption{Ablation studies on components of our model. (Scores are sorted by the delta from the full model.)}
\label{tab:ablation}
\end{table}

\noindent\textbf{Ablation Study}
We consider the contributions of several model components in \Cref{tab:ablation}.
The largest performance drop is from removing source-side copy,\footnote{All other hyper-parameter settings remain the same.}
showing its efficiency at reducing sparsity from open-class vocabulary entries.
Removing target-side copy also leads to a large drop.
Specifically, the subtask score of reentrancies drops down to 38.4\% when target-side copy is disabled.
Coverage loss is useful with regard to discouraging unnecessary repetitive nodes.
In addition, our model benefits from input features
such as language representations from BERT, index embeddings, 
POS tags,
anonymization indicators,
and character-level features from CharCNN.
Note that without BERT embeddings, our model still outperforms the previous best approaches~\citep{P18-1037,D18-1198} that are not using BERT.
Beam search, commonly used in machine translation,
is also helpful in our model.
We provide side-by-side examples in the Appendix to further illustrate the contribution from each component, which are largely intuitive, with the exception of BERT embeddings.  There the exact contribution of the component (qualitative, before/after ablation) stands out less: future work might consider a \emph{probing} analysis with manually constructed examples, in the spirit of \citet{Q16-1037,P18-1198,tenney2018iclr}.

In the last row, we only evaluate model performance at the edge prediction stage by forcing our model to 
decode the reference nodes at the node prediction stage. 
The results mean if our model could make perfect prediction at the node prediction stage,
the final \textsc{Smatch} score will be substantially high, which
identifies node prediction as the key to
future improvement of our model.

\begin{figure}[!ht]
\centering
\includegraphics[width=0.45\textwidth]{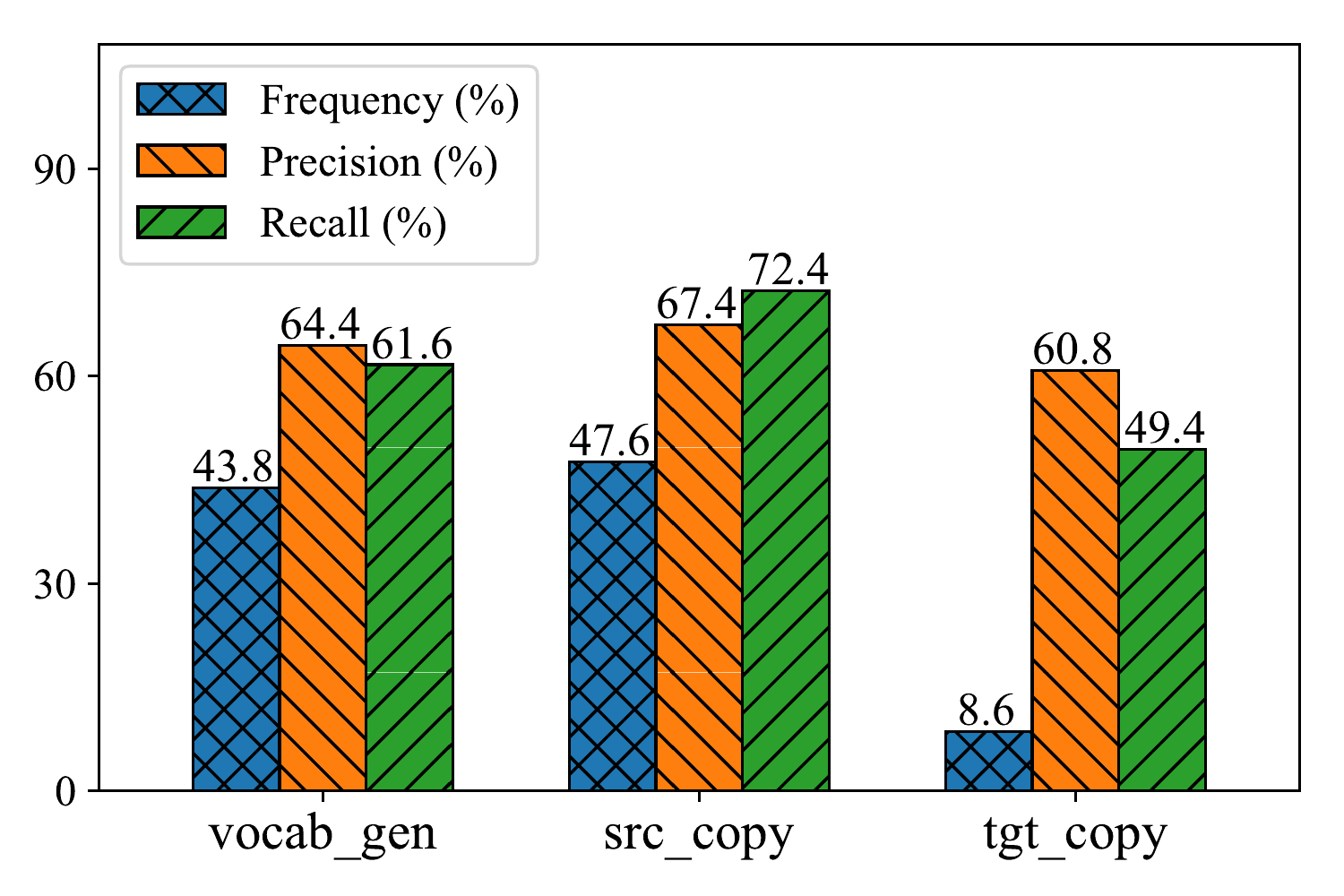}
\caption{Frequency, precision and recall of nodes from different sources,
based on the AMR 2.0 test set.}
\label{fig:freq-vs-f1}
\end{figure}

There are three sources for node prediction:
vocabulary generation, source-side copy, or target-side copy.
Let all reference nodes from source $z$ be $N^{(z)}_\text{ref}$,
and all system predicted nodes from $z$ be $N^{(z)}_\text{sys}$.
we compute frequency, precision and recall of nodes from source $z$ as below:
\begin{align*}
\left.\textrm{frequency}^{(z)} = {|N^{(z)}_\text{ref}|} \middle/ {\sum\nolimits_z |N^{(z)}_\text{ref}|}\right. \\
\left.\textrm{precision}^{(z)} = {|N^{(z)}_\text{ref}\cap N^{(z)}_\text{sys}|}\middle/{|N^{(z)}_\text{sys}|} \right.\\
\left.\textrm{recall}^{(z)} = {|N^{(z)}_\text{ref}\cap N^{(z)}_\text{sys}|}\middle/{|N^{(z)}_\text{ref}|}\right. 
\end{align*}

\Cref{fig:freq-vs-f1} shows the frequency of nodes from difference sources,
and their corresponding precision and recall based on our model prediction.
Among all reference nodes, 43.8\% are from vocabulary generation,
47.6\% from source-side copy, and only 8.6\% from target-side copy.
On one hand, the highest frequency of source-side copy helps address sparsity 
and results in the highest precision and recall.
On the other hand, we see space for improvement,
especially on the relatively low recall of target-side copy,
which is probably due to its low frequency.

\noindent\textbf{Node Linearization} As decribed in \Cref{sec:tech-overview}, 
we create the reference node list by
a pre-order traversal over the gold AMR tree.
As for the children of each node, we sort them in alphanumerical order.
This linearization strategy has two advantages: 
(1) pre-order traversal guarantees that a head node (\emph{predicate}) always comes in front of its children (\emph{arguments});
(2) alphanumerical sort orders according to role ID 
(i.e., \texttt{ARG0}\textgreater \texttt{ARG1}\textgreater ...\textgreater\texttt{ARGn}), 
 following intuition from research in Thematic Hierarchies~\cite{fillmore68:_case,levin2005argument}.

\begin{table}[!ht]
\centering
\begin{tabular}{@{}lcc@{}}
\toprule
Node Linearization   & \specialcell{AMR\\1.0} & \specialcell{AMR\\2.0} \\ \midrule
Pre-order + Alphanum  & 70.2   & 76.3    \\
Pre-order + Alignment      &  61.9  &  68.3  \\
Pure Alignment                 & 64.3  &   71.3   \\ \bottomrule
\end{tabular}
\caption{\textsc{Smatch} scores of full models trained and tested based on different node linearization strategies.}
\label{tab:order}
\end{table}

In \Cref{tab:order}, we report \textsc{Smatch} scores of full models trained and tested on data 
generated via our linearization strategy (Pre-order + Alphanum), as compared to two obvious alternates:
the first alternate still runs a pre-order traversal,
but it sorts the children of each node based on the their alignments to input words;
the second one linearizes nodes purely based alignments.
Alignments are created using the tool by~\citet{D14-1048}.
Clearly, our linearization strategy leads to much better results than the two alternates.
We also tried other traversal strategies such as combining in-order traversal with alphanumerical sorting or alignment-based sorting, but did not get scores even comparable to the two alternates.\footnote{
\citet{amr-seq2seq} also investigated linearization order, 
and found that alignment-based ordering yielded the best results under their setup 
where AMR parsing is treated as a sequence-to-sequence learning problem.}

\noindent\textbf{Average Pooling vs. Max Pooling}
In \Cref{fig:bert-encoder}, we apply average pooling to the outputs (last-layer hidden states) of BERT 
in order to generate word-level embeddings for the input sentence.
\Cref{tab:pooling} shows scores of models using different pooling functions.
Average pooling performs slightly better than max pooling.

\begin{table}[!ht]
\centering
\begin{tabular}{@{}lcc@{}}
\toprule
                & AMR 1.0 & AMR 2.0 \\ \midrule
Average Pooling & 70.2$\pm$0.1    & 76.3$\pm$0.1    \\
Max Pooling     & 70.0$\pm$0.1    & 76.2$\pm$0.1    \\ \bottomrule
\end{tabular}
\caption{\textsc{Smatch} scores based different pooling functions.
Standard deviation is over 3 runs on the test data.}
\label{tab:pooling}
\end{table}

\section{Conclusion}
We proposed an attention-based model for AMR parsing where 
we introduced a series of novel components into a transductive
setting that extend beyond what a typical NMT system would do on this task.
Our model achieves the best performance on two AMR corpora.
For future work, we would like to extend our model to
other semantic parsing tasks~\cite{oepen-etal-2014-semeval,ucca}.
We are also interested in semantic parsing in cross-lingual settings~\cite{x-dsp,damonte-cohen-2018-cross}.

\section*{Acknowledgments}
We thank the anonymous reviewers for their
valuable feedback. 
This work was supported in part by the
JHU Human Language Technology Center of Excellence (HLTCOE), and DARPA LORELEI and
AIDA. The U.S. Government is authorized to reproduce and distribute reprints for Governmental
purposes. The views and conclusions contained in
this publication are those of the authors and should
not be interpreted as representing official policies
or endorsements of DARPA or the U.S. Government.

\bibliography{acl2019}
\bibliographystyle{acl_natbib}

\appendix

\section{Appendices}
\label{sec:appendix}

\begin{figure*}[!ht]
\centering
\includegraphics[width=0.95\textwidth]{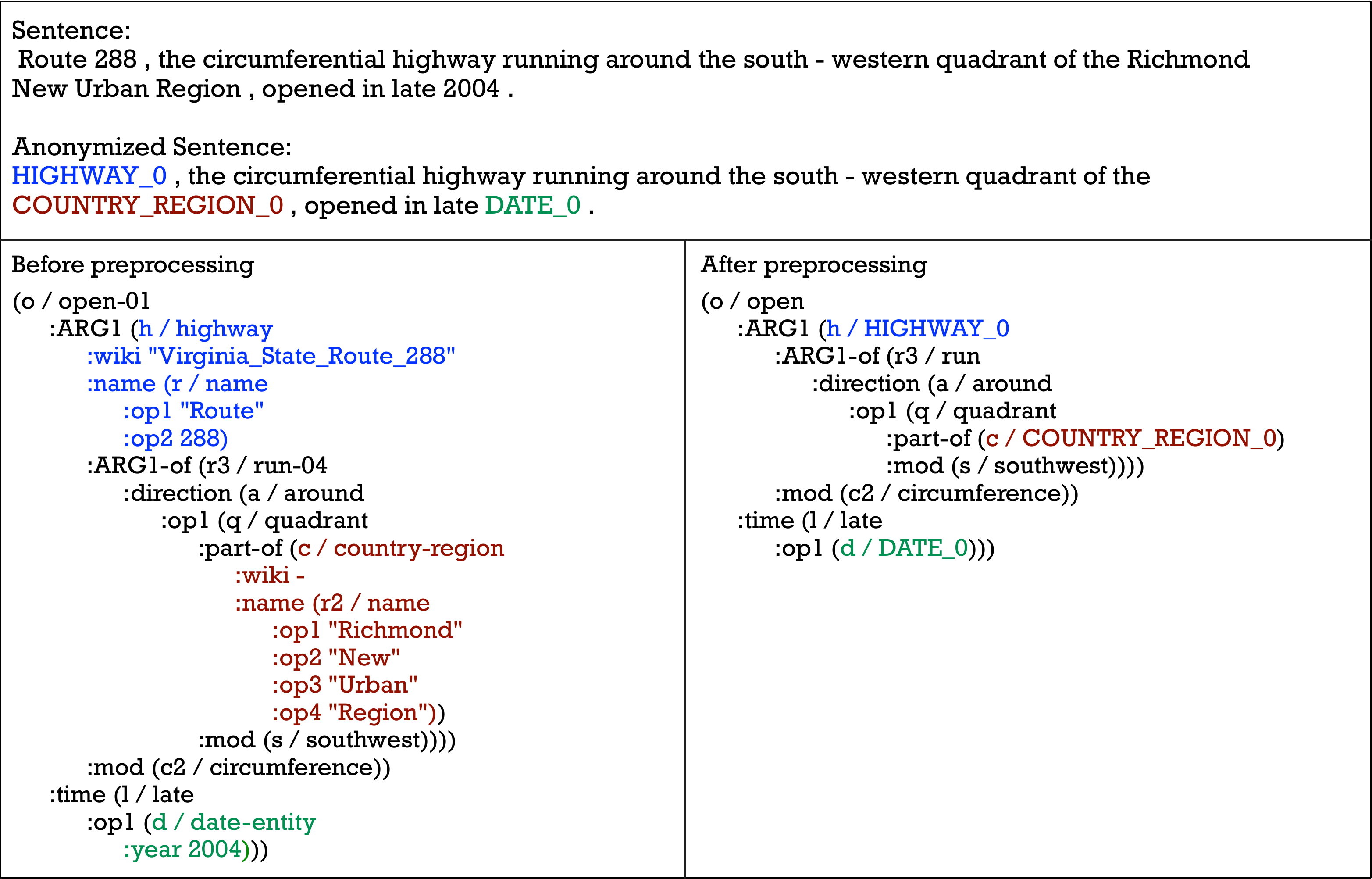}
\caption{An example AMR and the corresponding sentence before and after preprocessing. 
Senses are removed.
The first named entity is replaced by ``\textcolor{blue}{HIGHWAY\_0}''; 
the second named entity is replaced by ``\textcolor{cayenne}{COUNTRY\_REGION\_0}''; 
the first date entity replaced by ``\textcolor{moss}{DATE\_0}''.
\label{fig:preproc}}
\end{figure*}

\subsection{AMR Pre- and Post-processing}
Firstly, we to run Standford CoreNLP like \citet{P18-1037}, 
lemmatizing input sentences and adding POS tags to each token.
Secondly, we remove senses, wiki links and polarity attributes in AMR.
Thirdly, we anonymize sub-graphs of named entities and \texttt{*-entity}
in a way similar to~\citet{P17-1014}.
\Cref{fig:preproc} shows an example before and after preprocessing.
Sub-graphs of named entities are headed by one of AMR's fine-grained entity types
(e.g., \texttt{highway}, \texttt{country\_region} in \Cref{fig:preproc}) that
contain a \texttt{:name} role. 
Sub-graphs of other entities are headed by their corresponding entity type name
(e.g., \texttt{date-entity} in \Cref{fig:preproc}).
We replace these sub-graphs with a token of a special pattern ``TYPE\_i''
(e.g. \texttt{HIGHWAY\_0}, \texttt{DATE\_0} in \Cref{fig:preproc}), where
``TYPE" indicates the AMR entity type of the corresponding sub-graph,
and ``i'' indicates that it is the $i$-th occurrence of that type.
On the training set, we use simple rules to find mappings 
between anonymized sub-graphs and spans of text, and then replace mapped text with the
anonymized token we inserted into the AMR graph.
Additionally, we build a mapping of Standford CoreNLP NER tags
to AMR's fine-grained types based on the training set, which will be used in prediction.
At test time, we normalize sentences to match our anonymized training data.
For any entity span identified by Stanford CoreNLP, we replace it with a AMR 
entity type based on the mapping built during training.
If no entry is found in the mapping, we replace entity spans with the coarse-grained
NER tags from Stanford CoreNLP, which are also entity types in AMR.

In post-processing, we deterministically generate AMR sub-graphs for anonymizations 
using the corresponding text span.
We assign the most frequent sense for nodes (-01, if unseen) like~\citet{P18-1037}.
We add wiki links to named entities using the DBpedia Spotlight API~\cite{isem2013daiber}
following \citet{S16-1182,amr-seq2seq} with the confidence threshod at 0.5.
We add polarity attributes based on Algorithm \ref{alg:negation} where the four functions
\texttt{isNegation}, \texttt{modifiedWord}, \texttt{mappedNode}, and \texttt{addPolarity}
consists of simple rules observed from the training set.
We use the PENMANCodec\footnote{\url{https://github.com/goodmami/penman/}} to encode 
and decode both intermediate and final AMRs.

\begin{algorithm}[!ht]
\small
\SetAlCapNameFnt{\small}
\SetAlCapFnt{\small}
\SetKwFunction{MST}{maxArborescence}
\SetKwFunction{negation}{isNegation}
\SetKwFunction{head}{modifiedWord}
\SetKwFunction{node}{mappedNode}
\SetKwFunction{add}{addPolarity}
\SetKwProg{Fn}{Function}{}{}
\SetKwInOut{Input}{Input}\SetKwInOut{Output}{Output}
\Input{Sent. $\bm{w}=\langle w_1, ...,w_n\rangle$, Predicted AMR $A$
}
\Output{AMR with polarity attributes.}
\For{$w_i \in \bm{w}$}{
    \If{\negation{$w_i$}}{
        $w_j\leftarrow$ \head{$w_i$, $\bm{w}$}\; 
        $u_k\leftarrow$ \node{$w_j$, $A$}\;
        $A\leftarrow$ \add{$u_k$, $A$}\;
    }
}
\Return $A$\;
\caption{Adding polarity attributes to AMR.}\label{alg:negation}
\end{algorithm}

\subsection{Side-by-Side Examples}
In the next page, we provide examples from the test set, with side-by-side comparisons between
the full model prediction and the model prediction after ablation.

\begin{figure*}[!ht]
\centering
\includegraphics[width=0.95\textwidth]{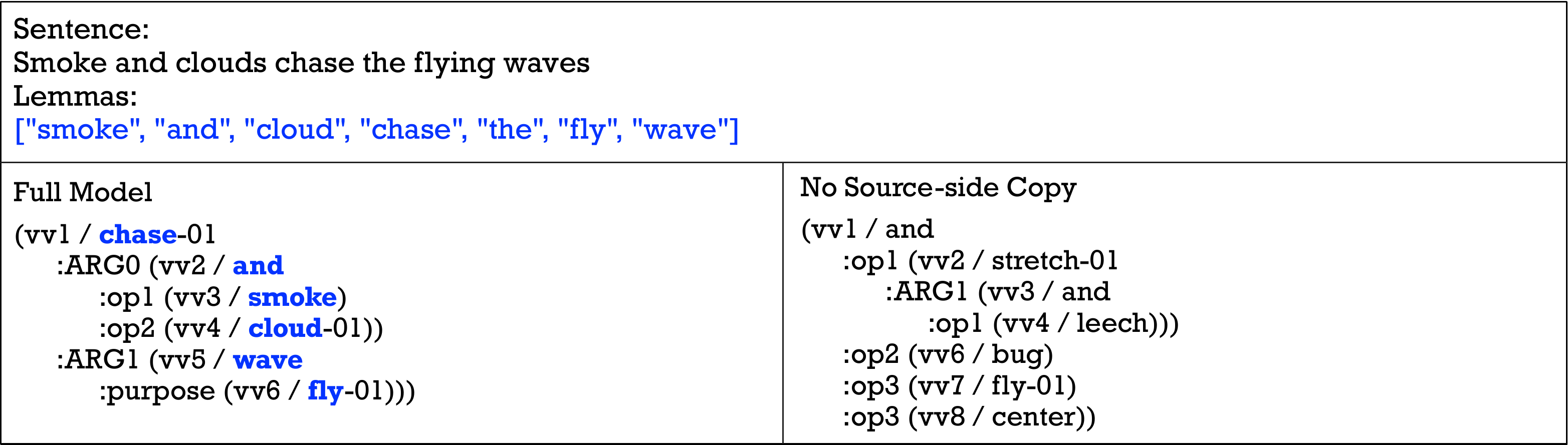}
\caption{Full model prediction vs. no source-side copy prediction. 
Tokens in blue are copied from the source side.
Without source-side copy, the prediction becomes totally different and inaccurate in this example.
\label{fig:no-source-copy-ex}}
\end{figure*}

\begin{figure*}[!ht]
\centering
\includegraphics[width=0.95\textwidth]{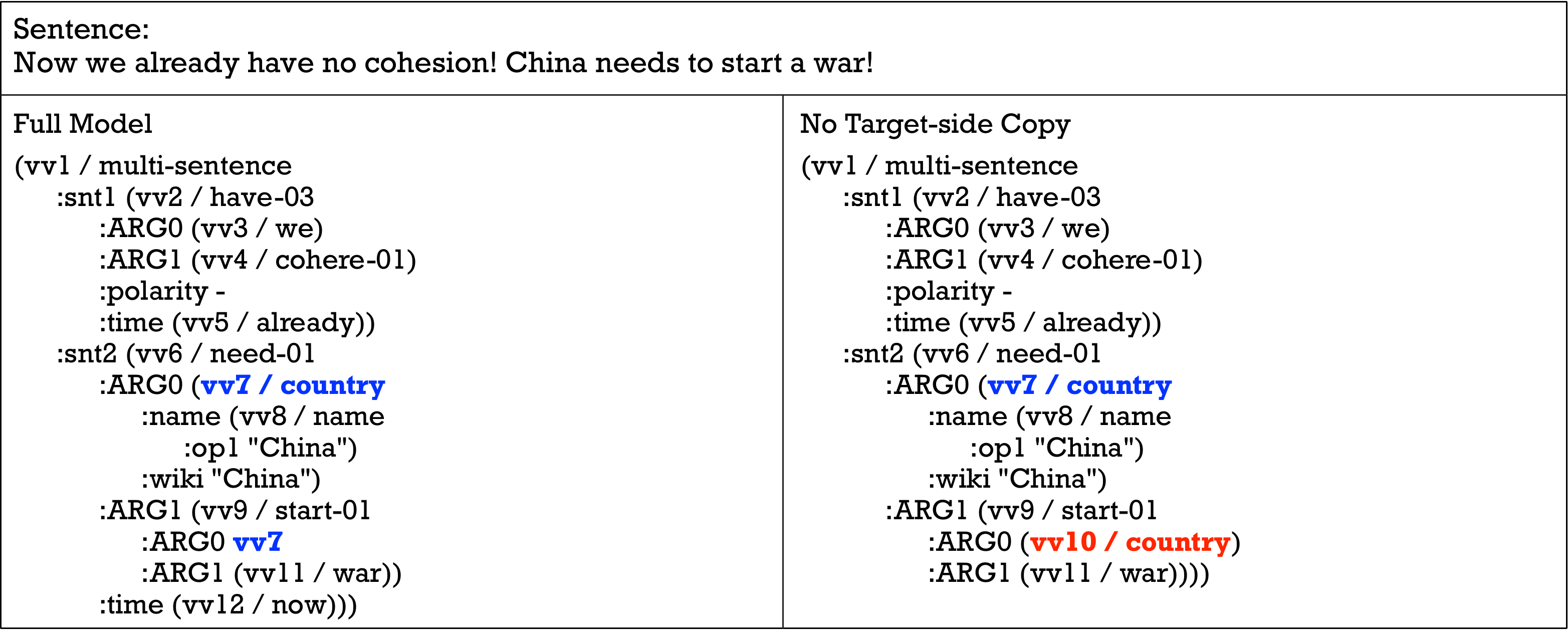}
\caption{Full model prediction vs. no target-side copy prediction. 
Nodes in blue denote the same concept (i.e., the country ``China'').
The full model correctly copies the first node
(``\textcolor{blue}{\text{vv7 / country}}'') as \texttt{ARG0} of ``\text{start-01}''.
Without target-side copy, the model has to generate a new node with a different index,
i.e., ``\textcolor{red}{\text{vv10 / country}}''.
\label{fig:no-target-copy-ex}}
\end{figure*}

\begin{figure*}[!ht]
\centering
\includegraphics[width=0.95\textwidth]{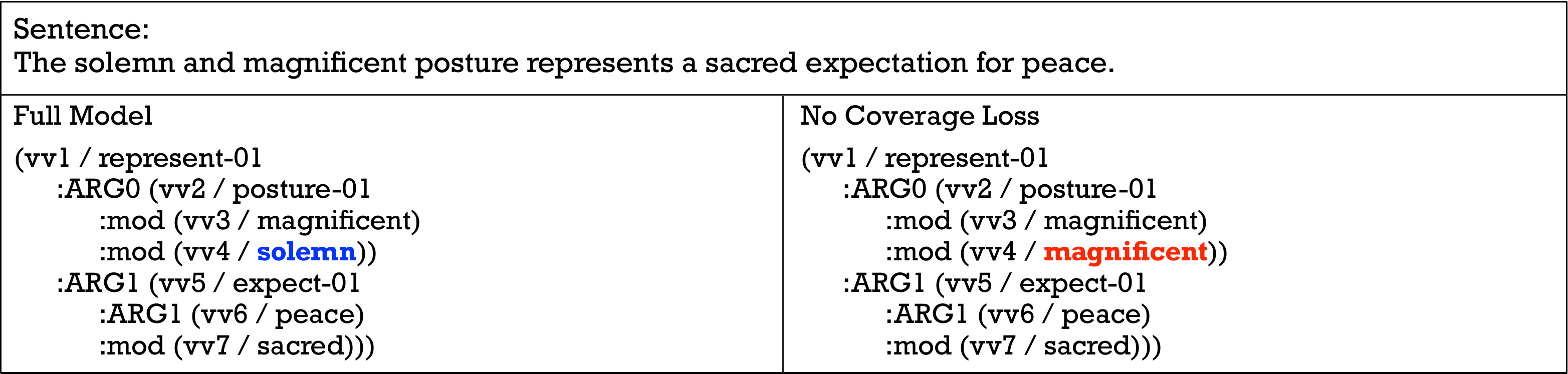}
\caption{Full model prediction vs. no coverage loss prediction. 
The full model correctly predicts the second modifier ``\textcolor{blue}{\text{solemn}}''.
Without coverage loss, the model generates a repetitive modifier ``\textcolor{blue}{\text{magnificent}}''.
\label{fig:no-coverage-ex}}
\end{figure*}

\begin{figure*}[!ht]
\centering
\includegraphics[width=0.95\textwidth]{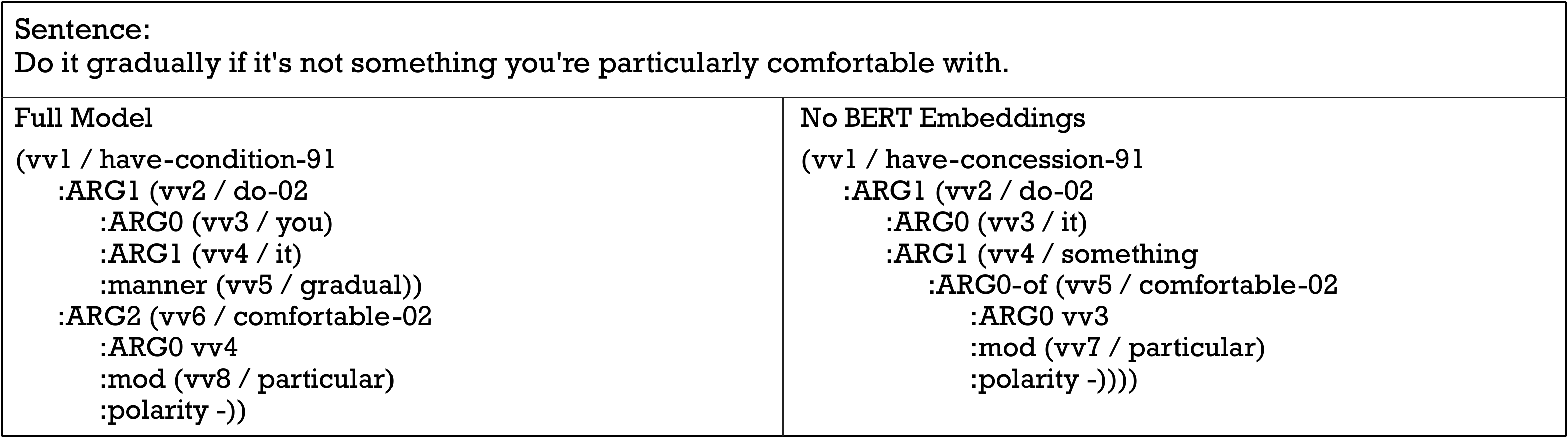}
\caption{Full model prediction vs. no BERT embeddings prediction. 
\label{fig:no-bert-ex}}
\end{figure*}

\end{document}